\documentclass[numsec,webpdf,contemporary,large]{oup-authoring-template}%
\graphicspath{{assets/}}

\usepackage{tikz}
\usetikzlibrary{arrows.meta}
\usepackage{makecell}
\usepackage{dsfont}

\usepackage{enumitem}
\usepackage{siunitx}
\usepackage{booktabs}
\usepackage{multirow}
\usepackage{graphicx}

\theoremstyle{thmstyleone}%
\theoremstyle{thmstyletwo}%
\theoremstyle{thmstylethree}%

\begin{document}

\journaltitle{Journal}
\DOI{-}
\copyrightyear{2023}
\pubyear{2023}
\access{}
\appnotes{Preprint}

\title[BioKGE]{Knowledge Graph Embeddings in the Biomedical Domain: Are They Useful?\\ A Look at Link Prediction, Rule Learning, and Downstream Polypharmacy Tasks}

\author[1,$\dagger$,$\ast$]{Aryo Pradipta Gema\ORCID{0009-0007-1163-3531}}
\author[1,$\dagger$,$\ast$]{Dominik Grabarczyk\ORCID{0009-0001-4340-430X}}
\author[1,$\dagger$,$\ast$]{Wolf De Wulf\ORCID{0000-0002-0219-3120}}
\author[1]{Piyush Borole\ORCID{0000-0003-3327-5847}}
\author[1,2,3]{Javier Antonio Alfaro\ORCID{0000-0002-5553-6991}}
\author[1,$\ddagger$]{Pasquale Minervini\ORCID{0000-0002-8442-602X}}
\author[1,$\ddagger$]{Antonio Vergari\ORCID{0000-0003-0036-5678}}
\author[1,$\ddagger$]{Ajitha Rajan\ORCID{0000-0003-3765-3075}}

\authormark{Gema et al.}
\address[1]{\orgdiv{School of Informatics}, \orgname{University of Edinburgh}, \orgaddress{\state{Edinburgh}, \country{United Kingdom}}}
\address[2]{\orgdiv{International Centre for Cancer Vaccine Science}, \orgname{University of Gda\'nsk}, \orgaddress{\state{Gda\'nsk}, \country{Poland}}}
\address[3]{\orgdiv{Department of Biochemistry and Microbiology}, \orgname{University of Victoria}, \orgaddress{\state{British Columbia}, \country{Canada}}}

\corresp[$\dagger$]{Joint first author.}
\corresp[$\ddagger$]{Joint senior authorship.}
\corresp[$\ast$]{Corresponding authors. \href{aryo.gema@ed.ac.uk}{aryo.gema@ed.ac.uk}, \href{s1873532@ed.ac.uk}{s1873532@ed.ac.uk}, \href{wolf.de.wulf@ed.ac.uk}{wolf.de.wulf@ed.ac.uk}, \href{arajan@ed.ac.uk}{arajan@exseed.ed.ac.uk}}


\abstract{
\textbf{Motivation:} 
Knowledge graphs are powerful tools for representing and organising complex biomedical data.
Several knowledge graph embedding algorithms have been proposed to learn from and complete knowledge graphs.
However, a recent study demonstrates the limited efficacy of these embedding algorithms when applied to biomedical knowledge graphs, raising the question of whether knowledge graph embeddings have limitations in biomedical settings.
This study aims to apply state-of-the-art knowledge graph embedding models in the context of a recent biomedical knowledge graph, BioKG, and evaluate their performance and potential downstream uses.\newline
\textbf{Results:}
We achieve a three-fold improvement in terms of performance based on the HITS@10 score over previous work on the same biomedical knowledge graph.
Additionally, we provide interpretable predictions through a rule-based method.
We demonstrate that knowledge graph embedding models are applicable in practice by evaluating the best-performing model on four tasks that represent real-life polypharmacy situations.
Results suggest that knowledge learnt from large biomedical knowledge graphs can be transferred to such downstream use cases.\newline
\textbf{Availability and implementation:}
Our code is available at \href{https://github.com/aryopg/biokge}{https://github.com/aryopg/biokge}.
}
\maketitle
\section{Introduction}\label{sec:intro}

Knowledge Graphs (KGs) are increasingly utilised for knowledge representation in the biomedical domain.
Recent studies show that KGs can be utilised to aid drug repurposing research~\cite{schultzMethodRationalSelection2021} and to predict the side effects of drug combinations~\cite{zitnikModelingPolypharmacySide2018, carlettiPredictingPolypharmacySide2021}.
To maximise the utility of KGs in this domain, comprehensive coverage of entities and links is essential.
A novel biomedical KG, called BioKG~\cite{walshBioKGKnowledgeGraph2020}, has been developed to be the first in the domain that attempts to agglomerate a wide range of entity and link types. However, accurately predicting links between entities in KGs can be challenging.

Knowledge Graph Embeddings (KGEs) offer a solution by representing KGs in a low-dimensional space~\cite{ferrariComprehensiveAnalysisKnowledge2022}.
However, the potential utility of KGEs in the biomedical field remains underexplored.
A recent study reports the limited success of KGEs for a biomedical KG~\cite{bonnerUnderstandingPerformanceKnowledge2022}, raising the question of whether KGE methods have reached their maximum efficacy in this domain. 

In this study, we show that it is possible to learn to accurately predict links in BioKG.
By taking into account recent studies outlining the best practices for KGE algorithms~\cite{ruffinelliYouCANTeach2020}, previously determined limitations can be overcome. 
Furthermore, the pretrained KGE models are also transferable to four downstream polypharmacy tasks, suggesting that a transfer learning paradigm where KGE models trained on large KGs are adapted for solving downstream tasks is feasible.
In addition, we investigate the efficacy of a rule-based model, called Anytime Bottom-Up Rule Learning~\cite[AnyBURL;][]{meilickeAnytimeBottomUpRule2019}, for BioKG. Such a rule-based model offers some degree of interpretability, which is important in the biomedical domain.

In summary, this study presents the following contributions:
\begin{itemize}[labelsep=3pt,itemsep=3pt,topsep=4pt]
    \item \textbf{A comprehensive evaluation of KGEs for BioKG using recent training best practices}, which reveals significant HITS@10 and mean reciprocal rank (MRR) improvement compared to previous results~\cite{bonnerUnderstandingPerformanceKnowledge2022}. The best-performing KGE model (ComplEx) reaches 0.793 HITS@10, compared to  0.286 in~\cite{bonnerUnderstandingPerformanceKnowledge2022}.
    \item \textbf{An investigation of the interpretability of a rule-based model for BioKG.} AnyBURL achieves a competitive HITS@10 score of 0.677 while providing interpretable rules.
    \item \textbf{An investigation of applying KGE models in real-world tasks.} The best-performing pretrained KGE model can easily be adapted to four downstream polypharmacy tasks in a transfer learning paradigm.
\end{itemize}

\section{Background}\label{sec:background}

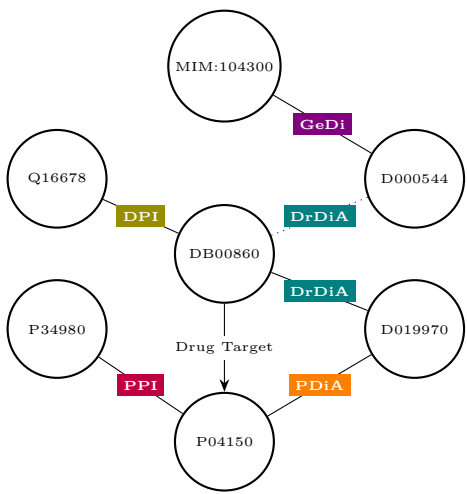
\begin{figure}[t]
    \centering
    \begin{tikzpicture}
    \tiny
\begin{scope}[every node/.style={circle,thick,draw}]
    \node[minimum size=1.3cm] (prednisolone) at (0,4) {DB00860};    
    \node[minimum size=1.3cm] (Alz) at (2.5,5)           {D000544};     
    \node[minimum size=1.3cm] (AlzG) at (0,6.5)        {MIM:104300}; 
    \node[minimum size=1.3cm] (C) at (0,1.5)             {P04150};  
    \node[minimum size=1.3cm] (D) at (2.5,3)             {D019970}; 
    \node[minimum size=1.3cm] (E) at (-2.2,5)            {Q16678}; 
    \node[minimum size=1.3cm] (F) at (-2.2,3)            {P34980}; 
\end{scope}

\begin{scope}[>={Stealth[black]},
              every node/.style={fill=white}]
    
    %
    %
    %
    %
    \path [->] (prednisolone) edge node{Drug Target} (C);
    \path [-] (prednisolone) edge node[white, fill=olive] {\textbf{DPI}} (E);
    \path [-] (prednisolone) edge node[white, fill=teal] {\textbf{DrDiA}} (D);
    \path [-] (C) edge node[white, fill=orange] {\textbf{PDiA}} (D);
    \path [-] (F) edge node[white, fill=purple] {\textbf{PPI}} (C);
    \path [-] (Alz) edge node[white, fill=violet] {\textbf{GeDi}} (AlzG);
    \path [dotted] (prednisolone) edge node[white, fill=teal] {\textbf{DrDiA}} (Alz);
\end{scope}
\end{tikzpicture}
    \caption{An extract of BioKG~\cite{walshBioKGKnowledgeGraph2020}. The nodes represent entities in the KG, edges between them are links. The variety in identifier structure shows that BioKG is a combination of multiple smaller KGs. In this extract, the centre node (DB00860) represents the drug \textit{prednisolone}, which targets the \emph{Glucocorticoid receptor} (P04150). This receptor is associated with disorders related to or resulting from the use of cocaine (D019970), indicated by the Protein-Disease-Association relation (PDiA). Hence, \emph{prednisolone} is connected to said disorders through the Drug-Disease-Association relation (DrDiA). The right-most node (D000544) represents \emph{Alzheimer's disease}, a genetic disorder (GeDi). One possible application that uses the information in the KG would be to train a model to predict missing links. Such a model could consider information from, for example, the Drug-Drug Interaction (DDI) and Protein-Protein Interaction (PPI) relations starting from \textit{prednisolone} to predict that \textit{prednisolone} could also be used to treat \emph{Alzheimer's disease}, as indicated by the dashed DrDiA relation between them.
    }
    \label{fig:biokg}
\end{figure}

\subsection{Link Prediction in Knowledge Graphs\label{sec:LP}}
KGs are a knowledge representation formalism in which knowledge about the world is modelled as relationships between entities~\cite{hoganKnowledgeGraphs2022}.
A KG can be represented as a set of subject-predicate-object triples, where each $(s, p, o)$ triple represents a relationship of type $p$ between the subject $s$ and the object $o$.
Link Prediction (LP) is the task of identifying missing triples, i.e. triples encoding true facts that are missing from the KG.
Consider, for example, the extract of BioKG presented in~\autoref{fig:biokg}. 
It contains information about \textit{prednisolone} (DB00860), a drug that targets the Glucocorticoid receptor. 
An LP model that is trained on BioKG could be used to fill in blanks in triples such as $(\text{DB00860}, \text{DrDiA}, \_)$, effectively predicting other disorders that \textit{prednisolone} could treat. 
Similarly, such a model could be used to fill in blanks in triples such as $(\_, \text{DrDiA}, \text{D000544})$, where D000544 is Alzheimer's disease, predicting drugs that could be repurposed to treat Alzheimer's. 
Recent work does this using a different biomedical KG, finding \textit{prednisolone} likely to be associated with Alzheimer's disease~\cite{nianMiningAlzheimerDiseases2022}. 
Moreover, early-stage investigations have confirmed that high doses of \textit{prednisolone} can result in some delay of cognitive decline~\cite{ricciarelliAmyloidCascadeHypothesis2017}.

Before considering such real-life applications, an LP model should be sufficiently evaluated using adequate baselines to avoid wasting resources in failed pharmaceutical trials.
An LP model's generalisation capabilities are evaluated using rank-based metrics.
To do so, the KG is partitioned into training, validation, and test triples.
For each test triple $(s,p,o)$, trained models are used to predict the subject or the tail, i.e. fill in blanks in $(\_,p,o)$ or $(s,p,\_)$, respectively. The resulting triples are then ranked based on how the model scores them.
Subsequently, triples aside from $(s,p,o)$ that exist in the training, validation, or test sets are filtered out such that other triples that are known to be true do not influence the ranking.
The resulting ranking is ultimately used to calculate metrics such as the MRR or the average HITS$@k$ (see~\autoref{app:metrics}).

\subsection{Knowledge Graph Embeddings}
A prevalent class of LP models come in the form of KGEs, which represent entities and relations as low-dimensional vectors and use a scoring function to indicate the plausibility of a triple.
Many models and training paradigms for embedding knowledge graphs have been proposed~\cite{wangKnowledgeGraphEmbedding2017,ferrariComprehensiveAnalysisKnowledge2022}.
Models usually differ in how entities and relation representations are used to compute the likelihood of a link in the KG. 
Examples are translational models such as TransE~\cite{bordesTranslatingEmbeddingsModeling2013}, TransH~\cite{wangKnowledgeGraphEmbedding2014}, and RotatE~\cite{sunRotatEKnowledgeGraph2019}, factorisation models such as DistMult~\cite{yangEmbeddingEntitiesRelations2015} and ComplEx~\cite{trouillonComplexEmbeddingsSimple2016}, and neural-network models such as ConvE~\cite{dettmersConvolutional2DKnowledge2018}. 
\autoref{tab:models} provides a summary of the domains in which these KGEs embed and the scoring functions they use.

\begin{table}[t]
    \centering
    \caption{
    An overview of KGE models, with the domain they embed in ($d$ corresponds to the embedding size) and their scoring function.
    Here, $*$ denotes the convolution operation,
    $\text{Re}(x)$ is the real part of $x \in \mathbb{C}$,
    $\langle \bm{x}, \bm{y}, \bm{z} \rangle = \sum_i \bm{x}_i \bm{y}_i \bm{z}_i$ denotes the tri-linear dot product, $g(x)$ is a nonlinear function, 
    $\text{vec}(x)$ is the flattening operator,
    $w$ denotes the convolutional filter,
    and $W$ denotes a linear transformation matrix.
    }\label{tab:models}
    \resizebox{\columnwidth}{!}{%
    \begin{tabular}{lcc}\toprule
         \bf Model    & \bf Domain & \bf Scoring function $f(\bm{e_s}, \bm{r_p}, \bm{e_o})$ \\ 
         \midrule
         TransE   \cite{bordesTranslatingEmbeddingsModeling2013} & $\mathbb{R}^d$   & $-\|\bm{e_s}+\bm{r_p}-\bm{e_o}\|$ \\
         TransH   \cite{wangKnowledgeGraphEmbedding2014} & $\mathbb{R}^d$   & $-\|(\bm{e_s} - \bm{w_p}^{\top} \bm{e_s} \bm{w_p}) +\bm{r_p} - (\bm{e_o} - \bm{w_p}^{\top} \bm{e_o} \bm{w_p})\|$ \\
         RotatE   \cite{sunRotatEKnowledgeGraph2019} & $\mathbb{C}^d$   & $-\|\bm{e_s}\circ\bm{r_p}-\bm{e_o}\|$\\ 
         DistMult \cite{yangEmbeddingEntitiesRelations2015} & $\mathbb{R}^d$   & $\langle \bm{e_s}, \bm{r_p}, \bm{e_o} \rangle$ \\
         ComplEx  \cite{trouillonComplexEmbeddingsSimple2016} & $\mathbb{C}^d$   & $\text{Re}(\langle \bm{e_s}, \bm{r_p}, \bm{e_o} \rangle)$\\ 
         ConvE    \cite{dettmersConvolutional2DKnowledge2018} & $\mathbb{R}^d$   & $g(\text{vec}(g([\bm{e_s},\bm{r_p}] * w))W)\bm{e_o}$\\\botrule 
    \end{tabular}
    }
\end{table}

The paradigms wherein these models are usually trained can vary in several ways, with free variables such as the loss function, regularisation, initialisation, and data augmentation strategies~\citep{ruffinelliYouCANTeach2020}.
%
%
Furthermore, KGs generally do not explicitly contain negative triples.
However, for a KGE to be trained in a way that allows it to differentiate between true and false triples, negative triples do need to be generated explicitly at training time.
Different approaches of generating negative samples consist of randomly corrupting certain selections of triples, possibly filtering out corrupted triples that already exist in the KG~\cite{bordesTranslatingEmbeddingsModeling2013,lacroixCanonicalTensorDecomposition2018,ruffinelliYouCANTeach2020}. 
Investigating the appropriate settings of these hyperparameters is important. 
While certain combinations of settings have been found to often significantly outperform others, the KG itself still dictates which settings are best~\cite{ruffinelliYouCANTeach2020}.

\subsection{Rule Learning\label{sec:RL}}
A different class of algorithms, called rule learning algorithms, predict links through logical rules extracted from the KG~\cite{meilickeAnytimeBottomUpRule2019}.
As an example, take the following rule:
\begin{align*}
    (\text{DB00860}, \text{DrDiA}, \text{D019970}) \rightarrow (\text{DB00860}, \text{DrDiA}, \text{D000544}),
\end{align*}
which states that if \textit{prednosolone} is a drug associated with cocaine-related disorders, it can also be a drug associated with Alzheimer's disease, simulating the LP example in a rule learning context. 
Such a rule may arise from frequently observed occurrences of drugs associated with both diseases and a lack of occurrences of drugs that only affect one of the two diseases.

A major advantage of rule learning algorithms is that, since predictions are based on concrete rules, one can provide explanations for the predictions.
Even if such rules are not easily understandable by humans, the effort to assess the likelihood that a prediction is sound might still be useful.
For example, in scenarios like drug discovery, where experimental confirmation is costly.
An argument often used against rule-based algorithms is that, when applied to large KGs, these systems would struggle with the exponentially increasing search space for possible rules, leading to a large computational overhead or incomplete rule-bases~\cite{zhangIterativelyLearningEmbeddings2019}.
Furthermore, rule learning models do not produce a latent representation of entities and relations.
Within the confines of a singular task, this is not a problem, but it means it is impossible to use these models as foundational models for downstream tasks.

\section{Evaluation Setup}\label{sec:setup}

\subsection{Dataset}\label{sec:dataset}

\subsubsection{BioKG}
Though a great number of biological databases exist, most are highly specialised in the kind of data they describe, focusing, for example, only on proteins \cite{auerDBpediaNucleusWeb2007, szklarczykSTRINGDatabase20112011} or drugs \cite{wishartDrugBankKnowledgebaseDrugs2008}.
Most attempts to adapt such data sources into KGs tailor them to specific projects. This limits possible downstream tasks compared to more generalist KGs.
BioKG minimises these limitations by combining knowledge from various databases, such as UniProt \cite{auerDBpediaNucleusWeb2007}, KEGG \cite{kanehisaKEGGReferenceResource2016}, and Reactome \cite{croftReactomeDatabaseReactions2011}. 
BioKG utilises metadata to provide valuable insights and predictions, focusing on a high-level analysis based on entity relations rather than on in-depth information, such as sequence data. While it also includes structural information, specifically through the MEMBER\_OF\_COMPLEX relation, its core strength lies in it providing a macroscopic view of complex biomedical data.

\begin{figure}
    \centering
    \includegraphics[width=\linewidth]{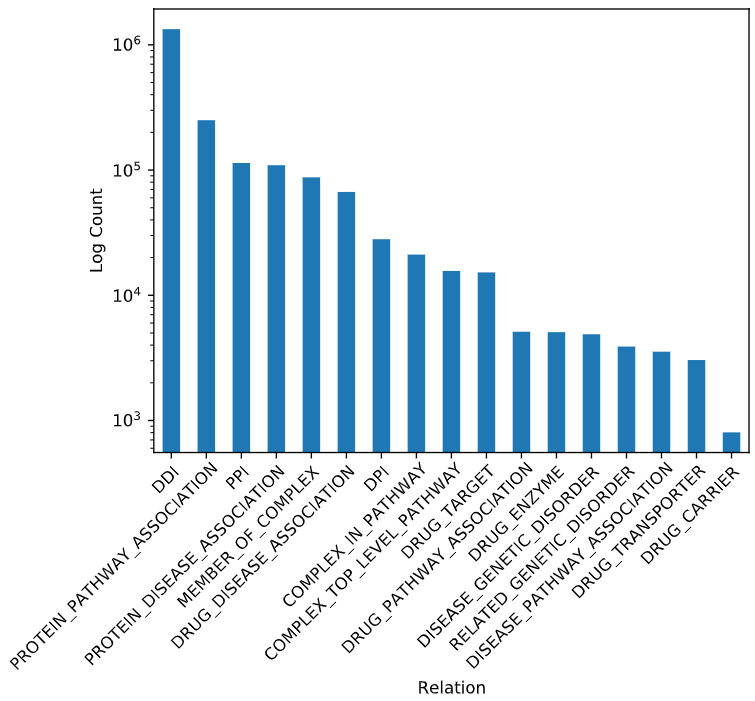}
    \caption{Frequencies of relationship types in BioKG on a logarithmic scale.}
    \label{fig:biokg_relationship_counts}
\end{figure}

In this work, we use BioKG version 1.0.0\footnote{\url{https://github.com/dsi-bdi/biokg/releases/tag/v1.0.0}} which contains 2,067,998 entries across 17 relations.
The number of entries per relationship varies extremely, as can be seen in~\autoref{fig:biokg_relationship_counts}.
An important metric of a KG is the degree of its nodes.
One recent study reasons that the in-degree has a large impact on how complex a model has to be to perform well on a KG~\cite{dettmersConvolutional2DKnowledge2018}.
A summary of these relationship statistics can be found in~\autoref{app:BioKGstats}.

A caveat of BioKG is that it does not explicitly track the directionality of relationships. 
Defining whether a relationship should be directed, however, is not a trivial task in the context of BioKG and requires a closer look at the data in question.
For example, drug-drug interaction (DDI) is a relationship between two entities of the same class, and trivially, if drug A interacts with B, then the reverse should hold and be explicitly modelled. With relationships denoting class membership, such as $(\text{cat}, \text{is\_a}, \text{mammal})$, the reverse does not hold.
However, for associative relationships between members of different categories, for example, drug-protein interaction (DPI), the matter is less clear. 
While the relationship is undirected, as the same protein will interact with the drug as well, these entries in BioKG always appear in the same order (drug before protein).
Modelling this as a directed edge might be sufficient, and adding the reverse might further imbalance the dataset. 
Therefore, reverses are only explicitly added in the cases of DDI and protein-protein interaction (PPI) since these are the only categories where the same entity class appears on both sides of the relationship, and all other categories have a consistent ordering in their subject and object classes.

\subsubsection{BioKG Polypharmacy Tasks}
Four smaller KGs that centre around the discovery of drug targets and the study of DDIs are published alongside BioKG. These KGs were constructed by building on pre-existing benchmark datasets and leveraging larger and more current KGs. The KGs are called DDI-Mineral, DDI-Efficacy, DPI-FDA, and DPI-FDA-EXP.

DDI-Mineral is centred around the study of DDIs and their association with abnormal mineral levels in the human body, with a particular emphasis on potassium, calcium, sodium, and glucose.
The KG is based on a previous KG~\cite{zitnikModelingPolypharmacySide2018}, which employs a dated TWOSIDES dataset~\cite{tatonettiDataDrivenPredictionDrug2012}. DDI-Mineral improves on this by integrating the more current and comprehensive DrugBank dataset.
It comprises 56,017 DDI triples across eight undirected relation types pertaining to 922 distinct drugs and their correlation with an increased or decreased likelihood of developing an abnormal mineral level.

DDI-Efficacy focuses on the relationship between DDIs and the therapeutic efficacy of the interacting drugs. It is similar to DDI-Mineral but with a specific focus on the polypharmacy side effects in relation to the efficacy of interacting drugs. It comprises 136,127 DDI triples across two undirected relation types that involve 3,342 unique drugs and their effect on the therapeutic efficacy of the interacting drugs, whether an increase or a decrease. Similar to DDI-Mineral, DDI-Efficacy is based on the DrugBank dataset.

DPI-FDA focuses on drug target protein interactions of FDA-approved drugs, compiled from the KEGG~\cite{kanehisaKEGGReferenceResource2016} and DrugBank databases. 
It contains 18,928 drug-protein interactions that involve 2,277 drugs and 2,654 proteins.
DPI-FDA is an extension of the previously published DrugBank\_FDA~\cite{wishartDrugBankKnowledgebaseDrugs2008} and Yamanishi09~\cite{yamanishiPredictionDrugTarget2008} KGS, with 9,881 triples and 5,127 triples, respectively.
In contrast, DPI-FDA is derived from more modern versions of the databases, leading to a more comprehensive and up-to-date representation of the drug target protein interactions of FDA-approved drugs.


DPI-FDA-EXP focuses on the effects of FDA-approved drugs on the protein expression levels in living systems. It comprises 903,429 triples with two directed relations describing the effects of 1,291 drugs on the expression of 55,196 proteins.

\subsection{Models}

\begin{figure*}
    \centering
    \includegraphics[width=\linewidth]{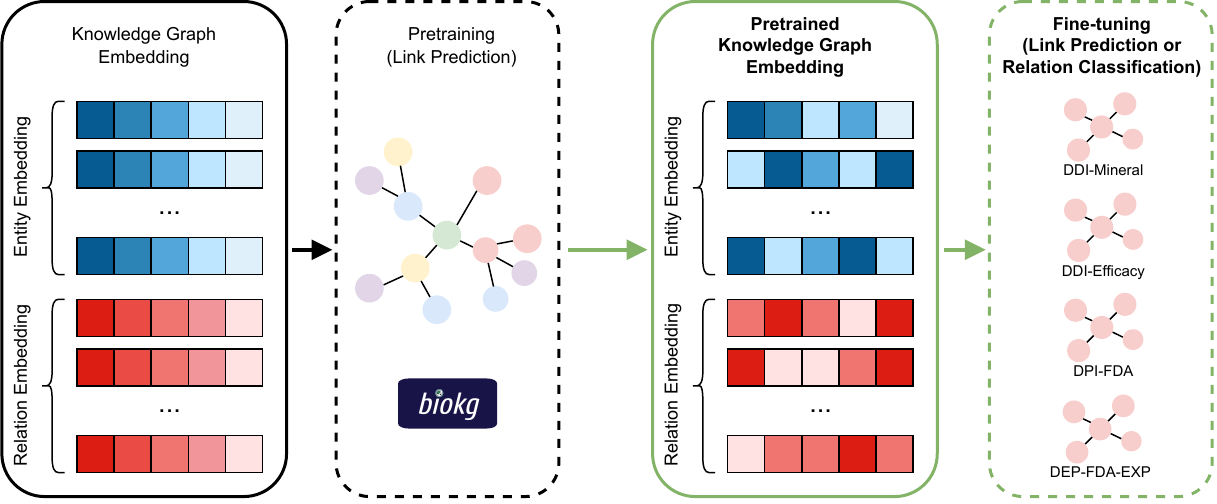}
    \caption{
    A visualisation of the KGE transfer learning framework from a large and all-encompassing KG, such as BioKG, to specific and purpose-driven downstream KGs.
    The KGE model is first pretrained using BioKG framed as an LP task. The pretrained LP model can then be utilised in more specific downstream KGs.
    In our study, we use the pretrained LP entity embedding and initialise the relation embedding from scratch as the downstream tasks introduce new types of relation.
    However, theoretically, one may use both pretrained LP entity and relation embedding should the downstream use cases contain a subset of the relation types in the larger KG.
    }
    \label{fig:biokge_framework}
\end{figure*}

\subsubsection{Link Prediction~\label{sec:models}}
We evaluate the six KGE models listed in~\autoref{tab:models}, as well as one rule-based model called AnyBURL.
These specific models were selected because they are widely recognised and commonly used in comparable studies~\cite{bonnerUnderstandingPerformanceKnowledge2022, ruffinelliYouCANTeach2020}.
Additionally, AnyBURL is included to provide a rule-based baseline for comparison against the KGE models.

All evaluations concerning the KGE models are performed using the \textit{LIBKGE} Python package~\cite{libkge}.
The hyperparameters of the KGE models are optimised on the validation split through 30 quasi-random hyperparameter optimisation (HPO) trials.
The negative sampling method and loss function are always fixed. Previous work on KGs with statistics similar to those of BioKG has shown that \textit{1vsAll} and CE loss are often the best choices for the selection of models that are implemented in this work~\cite{ruffinelliYouCANTeach2020}. However, this combination is computationally infeasible for RotatE and TransH. Whenever \textit{1vsAll} was found to be infeasible, negative sampling is used.
A table that contains all the hyperparameters and the ranges over which they are optimised can be found in \autoref{app:hyperparameters}. Seeded \textit{LIBKGE} configuration files and the training, validation, and test splits (80/10/10) that allow reproducing all the HPO runs can be found on the accompanied code repository.

To assess the value of transfer learning from BioKG to downstream applications, the best-performing LP model is also evaluated on the BioKG polypharmacy tasks, referred to as the pretrained LP model.
We compare this pretrained LP model to one trained from scratch in terms of the HITS@10 and MRR scores as well as the number of epochs required to reach the peak performance.
\autoref{fig:biokge_framework} visualises this transfer learning framework.

AnyBURL is trained using the original Java implementation\footnote{\url{https://web.informatik.uni-mannheim.de/AnyBURL/}} at version 23-1. One key difference with the KGE models is that no negative sampling is necessary and that there is no loss function. Instead, the hyperparameters are limited to the training time, the confidence threshold to retain rules, and the maximum rule length. Optimal training time depends significantly on the number of allocated CPUs; in the case of this work, 76 CPUs were used. HPO is performed by grid search, with training times between 100 and 1000 in intervals of 100, thresholds of 0.1, 1 and 10 and rule lengths of 1, 2, 3, and 4. The source code for the AnyBURL evaluation can be found on a separate repository\footnote{\url{https://github.com/Dominko/biokg_anyburl}}


\subsubsection{Relation Prediction}
In addition to LP, the BioKG polypharmacy tasks can be further analysed as multiclass relation classification tasks, where the focus specifically lies on predicting the correct relations given pairs of entities as inputs.
In such a context, the task KGs require an additional relation type that denotes no interaction between the entities to make the tasks more realistic. In practice, the model receives a pair of entities, such as `Nebivolol' and `Lofexidin', and predicts the corresponding relation. For instance, in DDI-Efficacy, the model should predict either `decrease therapeutic efficacy'', `increase therapeutic efficacy', or `no effect'.

The objective of this study is to evaluate the performance of a classification model that uses a pretrained LP embedding taken from the best-performing pretrained LP model in BioKG compared to a model that is initialised from scratch.
To do so, an entity embedding layer is extracted from the pretrained LP model that performs best on BioKG, with the addition of a softmax output layer.
The baseline setup has an identical network architecture and uses the same hyperparameters but does not have the pretrained LP embedding weights.
\autoref{fig:biokge_framework} visualises this transfer learning framework to downstream tasks framed as relation prediction tasks.
The goal is to verify the hypothesis that the model with the pretrained LP embedding outperforms the baseline model.
Additionally, the performance of the same model with pretrained and frozen LP embedding is compared to the same model with pretrained and fine-tuned LP embedding. During training, all the models are configured to use an embedding vector of size 512, a batch size of 512, and a learning rate of \num{1e-4}.
The model performance is evaluated with standard classification metrics, i.e. Area Under the Receiver Operating Characteristic curve (AUROC), Area Under the Precision-Recall curve (AUPRC), and Mean Average Precision (MAP).

\section{Results \& Analyses}\label{sec:results}

\subsection{Link Prediction in BioKG}
\autoref{tab:lp-results} presents the LP results for BioKG for each of the best-found configurations of the models.
The complete configurations of the KGE models can be found in \autoref{tab:config} in \autoref{app:config}.
In terms of MRR and HITS@10, ComplEx performs best. Generally, the results suggest that factorisation models such as ComplEx and DistMult outperform other classes of models. Translational models perform worst. This indicates that similarity-based scoring functions might be a good fit for biomedical KGs such as BioKG. An interesting observation is that the best DistMult configuration is able to achieve a relatively high HITS@10 of 0.667 and MRR of 0.471 in a relatively low number of epochs (14) and with a relatively small embedding size (128). More extensive HPO might allow DistMult to achieve scores similar to those of ComplEx. If so, this would suggest that embedding in complex space instead of in Euclidean space is not a requirement for performing well on BioKG. ConvE performs similarly to ComplEx, indicating that a substantial amount of information can be captured from a triple's local neighbourhood~\cite{dettmersConvolutional2DKnowledge2018}.

As a baseline for the KGE models, a comparison is made to results in the literature. Specifically, those of a recent study that performs similar evaluations on the same version and splits of BioKG~\cite{bonnerUnderstandingPerformanceKnowledge2022}.
The authors perform HPO runs for all KGE models except for ConvE.
The results presented in \autoref{tab:lp-results} are all significantly better than what is reported in the previous study~\cite{bonnerUnderstandingPerformanceKnowledge2022}. For ComplEx, the aforementioned study reports a HITS@10 of $0.012$, compared to $0.793$ here.
For DistMult, this is $0.082$, compared to $0.667$ here. The best-performing model in the previous study is RotatE, which achieved a HITS@10 of $0.286$, compared to $0.618$ here, and compared to $0.793$ of the best-performing model here (ComplEx). TransE and TransH also achieve higher scores, with HITS@10 of $0.239$ and $0.080$ compared to $0.474$ and $0.574$ here, respectively. Since the data and splits are all the same, an explanation for these results must lie elsewhere.
The only differences with the evaluations in~\cite{bonnerUnderstandingPerformanceKnowledge2022} are the way negative samples are generated and the loss function. In~\cite{bonnerUnderstandingPerformanceKnowledge2022}, negative sampling and margin ranking loss are implemented for all models.
Based on the evaluations on KGs with statistics similar to those of BioKG in another work~\cite{ruffinelliYouCANTeach2020}, 1vsAll (where feasible, negative sampling otherwise, see Section~\ref{sec:models}) and CE loss seemed like better choices. The presented results confirm that this is the case, which underlines the importance of optimal training setup.

\begin{table}[t]
\centering
\caption{LP performance of KGE models and one rule-based model on BioKG. Each entry corresponds to the best configuration found in 30 quasi-random HPO trials. }\label{tab:lp-results}
\begin{tabular}{@{}lccccc@{}}
    \toprule
    \multirow{2}{*}{\bf Model} & \multirow{2}{*}{\bf Epochs} & \multirow{2}{*}{\bf Emb. Size} & \multirow{2}{*}{\bf MRR} & \multicolumn{2}{c}{\bf HITS@10} \\
    \cmidrule(lr){5-6}
            & & & & This work & \cite{bonnerUnderstandingPerformanceKnowledge2022}$^{**}$ \\ \midrule
    ComplEx      & 184     & 512       & 0.629  & 0.793 & 0.012  \\ 
    DistMult     & 14      & 128       & 0.471  & 0.667 & 0.082  \\ 
    TransE       & 124     & 256       & 0.273  & 0.474 & 0.239  \\ 
    TransH       & 200     & 256       & 0.280  & 0.574 & 0.080  \\ 
    RotatE       & 200     & 1024      & 0.422  & 0.618 & 0.286  \\ 
    ConvE        & 94      & 1024      & 0.599  & 0.765 & --     \\ 
    AnyBURL      & 200s*   & --        & 0.557  & 0.678 & --     \\ \botrule
    \multicolumn{6}{l}{$^*$ AnyBURL training time is in seconds.}\\    \multicolumn{6}{l}{$^{**}$ Average over their five best HPO trials.}
\end{tabular}
\end{table}

\subsection{Rule Learning}

With the standard setup, AnyBURL achieves HITS@10 similar to that of DistMult.
The hyperparameter optimisation of AnyBURL does not yield significant differences.
The threshold has no notable impact on the Hits@10 and MRR.
Increasing rule length by 1, makes the runtime, memory, and storage requirements prohibitively high.
It should also be noted that AnyBURL explicitly separates learning and inference. Hits@10 is close to the peak after as little as 200 seconds, but inference takes up to 12 hours with 76 CPUs.

Nevertheless, AnyBURL predictions can be connected back to the rules that generate them, and thus provide explanations for why they have been made. Consider the examples generated using AnyBURL below:
\begin{align}
(x, \text{DrDiA}, \text{D006099}) &\rightarrow (x, \text{DrDiA}, \text{D006973})\label{rule:1}\\
(x, \text{DrDiA}, \text{D013610}) &\rightarrow (x, \text{DrDiA}, \text{D007022})\label{rule:2}
\end{align}
Rule (\ref{rule:1}) states that if drug $x$ is associated with \textit{granuloma} (D006099), then it will also be related to \textit{hypertension} (D006973). 
Similarly, rule (\ref{rule:2}) states that if drug $x$ interacts with \textit{Tachycardia} (D013610), it will also interact with \textit{hypotension} (D007022.) 
AnyBURL also supplies a confidence score in the form of the total number of occurrences of the right-hand side and co-occurrences of both. 
For example, for the rule (\ref{rule:2}), there were 238 occurrences and 150 co-occurrences in the training data, resulting in a confidence score of 63\%.
These might be interesting in contexts such as drug discovery, where specialists can interpret such statistical connections to assess their likelihood in the real world.

\subsection{Adapting to Downstream Polypharmacy Tasks}
We use the best-performing model for BioKG, ComplEx, and evaluate it on the BioKG polypharmacy KGs.
\autoref{tab:lp-bench-results} presents the results for all four polypharmacy KGs using ComplEx in two different configurations - (1) when CompleEx is trained from scratch, and (2) when it is initialised with the embeddings of the configuration of ComplEx that performed best on BioKG (denoted as ComplEx-P). The generation of negative triples and the loss function are kept fixed as 1vsAll and CE, respectively. To use the pretrained LP embeddings, the embedding size is fixed. To improve comparability, this is also done for the instance of ComplEx trained from scratch.

\begin{table}[t]
\centering
\caption{Performance of ComplEx and ComplEx-P on the BioKG polypharmacy KGs as LP tasks, where ``-P'' indicates that ComplEx was initialised with pretrained LP embeddings. Each entry corresponds to the best configuration found in 30 quasi-random HPO trials. The generation of negative samples and the loss function are fixed as 1vsAll and CE, respectively. The embedding size is fixed at 512, such that the learnt embeddings of the best ComplEx configuration from \autoref{tab:lp-results} can be used as initialisation.}\label{tab:lp-bench-results}
\begin{tabular}{@{}llccc@{}}
\toprule
\bf Benchmark     & \bf Model          & \bf Epochs & \bf HITS@10 & \bf MRR   \\ \midrule
DDI-Efficacy & ComplEx        & 196    & 0.962   & 0.847 \\
              & ComplEx-P      & 52     & 0.975   & 0.865 \\ \midrule
DDI-Minerals & ComplEx        & 106    & 0.976   & 0.861 \\
              & ComplEx-P      & 54     & 0.987   & 0.884 \\ \midrule
DPI-FDA      & ComplEx        & 76     & 0.549   & 0.386 \\
              & ComplEx-P      & 4      & 0.742   & 0.542 \\ \midrule
DEP-FDA-EXP & ComplEx        & 72     & 0.287   & 0.171 \\
              & ComplEx-P      & 22     & 0.304   & 0.185 \\ \botrule
\end{tabular}
\end{table}

Both models perform very well for DDI-Efficacy and DDI-Minerals.
A HITS@10 in the 90s and MRR in the 80s indicates that ComplEx is able to accurately rank general triples and specific triples. Performance for the DPI-FDA and DEP-FDA-EXP KGs lies lower. It must be noted that DPI-FDA is a significantly smaller KG compared to the others, whereas DEP-FDA-EXP is a significantly larger KG. Hence, overfitting and underfitting, respectively, must be considered as causes of these results. In general, these two benchmarks seem to be relatively harder.

When comparing ComplEx to ComplEx-P, two observations can be made.
Firstly, ComplEx-P's best-found configuration requires significantly fewer epochs to reach similar or better performance for all four benchmarks. Secondly, for DPI-FDA and DEP-FDA-EXP, the best configuration for ComplEx-P produces more accurate results in comparison with the best configuration for ComplEx. 
The obvious decrease in epochs required for ComplEx-P to reach performance similar to that of ComplEx strongly suggests that the pretrained LP embeddings, in the least, allow for a positive transfer of knowledge in the form of a good initialisation for these tasks.
The case of DPI-FDA is particularly interesting because it resembles a common scenario in the biomedical world where there is little training data. The results show that performance and training time in these scenarios can be improved upon by pretraining using larger KGs.

\autoref{tab:rc-bench-results} presents the results for all four tasks when the focus lies on predicting the relation only.
For DDI-Efficacy, DDI-Minerals, and DPI-FDA, all models demonstrate near-perfect performance in all classification metrics.
This is not the case for DEP-FDA-EXP, indicating that this task is more complex than the others.
The pretrained and fine-tuned model shows the best AUROC and MAP scores, with 0.908 and 0.759, respectively.
The pretrained but frozen model shows the best AUPRC score of 0.812.
These results suggest that the polypharmacy tasks are trivial when framed as relation classification tasks.
Consequently, no conclusions can be drawn regarding knowledge transfer in this setting.
Further study should look into more difficult tasks to verify whether models with pretrained LP embedding weights can perform better than models that are trained from scratch.

\begin{table}[t]
\centering
\caption{Classifier performance on the BioKG polypharmacy KGs as relation prediction tasks. In the model column, ``NN'' denotes a fully-connected feed-forward neural network, ``-PF'' denotes a pretrained and frozen embedding, and ``-P'' denotes a pretrained and fine-tuned embedding.}\label{tab:rc-bench-results}
\begin{tabular}{@{}llcccc@{}}
\toprule
\bf Benchmark     & \bf Model & \bf AUROC & \bf AUPRC & \bf MAP   \\ \midrule
DDI-Efficacy & NN    & 0.970 & 0.936 & 0.935 \\
              & NN-PF & 0.968 & 0.933 & 0.933 \\
              & NN-P  & 0.970 & 0.944 & 0.938 \\ \midrule
DDI-Minerals & NN    & 0.998 & 0.995 & 0.995 \\
              & NN-PF & 0.997 & 0.986 & 0.986 \\
              & NN-P  & 0.998 & 0.995 & 0.995 \\ \midrule
DPI-FDA      & NN    & 1     & 1     & 1     \\
              & NN-PF & 1     & 1     & 1     \\
              & NN-P  & 1     & 1     & 1     \\ \midrule
DEP-FDA-EXP & NN    & 0.889 & 0.791 & 0.738 \\
              & NN-PF & 0.908 & 0.761 & 0.759 \\
              & NN-P  & 0.891 & 0.812 & 0.739 \\ \botrule
\end{tabular}
\end{table}

\section{Conclusion}\label{sec:conc}

Various models are evaluated for performing LP and relation classification on a biomedical KG, i.e. BioKG~\cite{walshBioKGKnowledgeGraph2020}.
The presented results show that the performance of the models are superior compared to those reported in a recent similar study~\cite{bonnerUnderstandingPerformanceKnowledge2022}.
The results suggest that factorisation models such as ComplEx and DistMult and neural network models such as ConvE generally perform better than the other classes of models. The relative success of ConvE suggests that further investigation into neural network models could be fruitful. Graph Neural Networks (GNN), also known to capture local information~\cite{zhouGraphNeuralNetworks2020}, could be interesting to evaluate.

ComplEx achieves HITS@10 of 0.793 and MRR of 0.629 on BioKG, followed by ConvE with 0.765 HITS@10 and 0.599 MRR.
As the best-performing model on BioKG, ComplEx is evaluated on the BioKG benchmarks.
The evaluations compare ComplEx when trained from scratch with ComplEx when initialised with the embeddings of the configuration of ComplEx that performed best on BioKG.
The comparison shows that pretrained ComplEx requires significantly fewer epochs to outperform ComplEx without pretraining.
Particularly, in a task with little training data (DPI-FDA), pretrained ComplEx significantly outperforms its non-pretrained counterpart, with HITS@10 of 0.742 compared to 0.549, respectively.

Though the rule learning model AnyBURL does not achieve as high a performance as the best-performing embedding models, its extremely short training time and explainable rules make it an interesting avenue for future research. Particularly, increasing the rule length might yield a more competitive performance, though performance and memory bottlenecks need to be overcome, by, for example, creating approaches using GPUs rather than CPUs or better rule aggregation methods, such as SAFRAN~\cite{ottSAFRANInterpretableRulebased2021}.

An additional experiment frames the BioKG benchmark tasks as relation classification tasks.
Similar to the LP benchmarking setting, the best-performing ComplEx configuration is utilised further in this experiment.
The entity embedding that ComplEx learns for BioKG is extracted to be used as features in classification tasks.
The results show that the BioKG benchmark tasks are trivial when framed as relation classification tasks.
A simple feed-forward NN model can achieve near-perfect results, and pretrained LP embeddings provide only marginal improvements in performance in the DEP-FDA-EXP task, with 0.908 AUROC compared to 0.889.
Future studies should analyse the triviality of the downstream polypharmacy tasks more closely. Although we reach near-perfect performance here, the task of predicting the relation between these entities is generally considered to be hard~\cite{bansalCommunityComputationalChallenge2014, tatonettiDataDrivenPredictionDrug2012}. Future work should evaluate the usefulness of pretrained LP embedding weights on challenging tasks that require multi-hop reasoning with different entity types.



In summary, this study emphasises the potential utility of KGEs for predicting yet-to-be-known interactions between biomedical entities.
This capability has the potential to reduce costs in biomedical research, such as optimising the selection of candidate compounds in drug research.
This study also investigates an interpretable rule-based model that performs comparably and may be of interest in biomedical research that prioritises explainability.
Furthermore, this study demonstrates that knowledge in KGEs is transferable from large and comprehensive KGs such as BioKG \cite{walshBioKGKnowledgeGraph2020} to data-sparse domains such as polypharmacy prediction.
The results from four downstream polypharmacy tasks highlight the feasibility of implementing such approaches in scenarios where data collection is expensive, which are exactly the domains in which effective candidate selection carries the largest benefits.




\section{Competing interests}
No competing interest is declared.

\section{Author contributions statement}

A.P.G. conceived the general experiments planning. A.P.G. worked on relation prediction.
D.G. worked on the rule-learning experiments.
W.D.W. worked on the knowledge graph embedding models for the link prediction experiments.
A.P.G., D.G., and W.D.W. analysed and interpreted the results and wrote the manuscript jointly. P.B. and J.A.A. helped with understanding the KG entities and relations. 
A.R., A.V. and P.M. supervised the project and reviewed the manuscript.

\section{Acknowledgements}
This work was supported by the United Kingdom Research and Innovation (grant EP/S02431X/1), UKRI Centre for Doctoral Training in Biomedical AI at the University of Edinburgh, School of Informatics. For the purpose of open access, the author has applied a Creative Commons attribution (CC BY) licence to any author-accepted manuscript version arising.

This work made use of the resources provided by the Edinburgh Compute and Data Facility (ECDF, \url{www.ecdf.ed.ac.uk/}), as well as those provided by the Cambridge Service for Data-Driven Discovery (CSD3) operated by the University of Cambridge Research Computing Service (\url{www.csd3.cam.ac.uk}), provided by Dell EMC and Intel using Tier-2 funding from the Engineering and Physical Sciences Research Council (capital grant EP/T022159/1), and DiRAC funding from the Science and Technology Facilities Council (\url{www.dirac.ac.uk}).

P.M. was partially funded by the European Union’s Horizon 2020 research and innovation programme under grant agreement no. 875160, the Edinburgh Laboratory for Integrated Artificial Intelligence (ELIAI, \url{web.inf.ed.ac.uk/eliai/}) EPSRC (grant no. EP/W002876/1), an industry grant from Cisco, and a donation from Accenture LLP. 

A.R., J.A.A., and P.B. are funded by European Union's Horizon 2020 research and innovation programme under grant agreement No. 101017453, with A.R. also being funded by a Royal Society Industry Fellowship. 

\bibliographystyle{plain}
\bibliography{reference}

\onecolumn
\begin{appendices}

\section{Evaluation Metrics\label{app:metrics}}
We give a formal definition of the evaluation metrics used in this work.
If $\mathcal{E}$ is the set of all entities in the KG and $\mathcal{K}^{\text{test}}$ is the test set of triples, denote for a given triple $(s,p,o)$:
\begin{itemize}[labelsep=4pt]
    \item The set of filtered pseudo-negative triples:
    \begin{align*}
        \{(s,p,o') :t' \in \mathcal{E} \text{ and } (s,p,o') \text{ does not appear in the training, validation or test triples}\}
    \end{align*}
    \item score$(s,p,o)$ the model's score for $(s,p,o)$.
    \item rank$(o\mid s,p)$ (and symmetrically rank$(s\mid p,o)$) the filtered rank of entity $t$, i.e. the rank of score$(s,p,o)$ in the scores of all triples in the set of filtered pseudo-negative triples. For ties in scores, we take the mean rank of all triples with score$(s,p,o)$.
\end{itemize}
Then the evaluation metrics are defined as:
\begin{align*}
    \text{MRR} &= \frac{1}{2|\mathcal{K}^{\text{test}}|} \sum\limits_{(s,p,o)\in \mathcal{K}^{\text{test}}} \left(\frac{1}{\text{rank}(s \mid p,o)} + \frac{1}{\text{rank}(o \mid s,p)}\right),\\
    \text{HITS@k} &= \frac{1}{2|\mathcal{K}^{\text{test}}|} \sum\limits_{(s,p,o)\in \mathcal{K}^{\text{test}}} \left(\mathds{1}(\text{rank}(s \mid p,o) \leq k) + \mathds{1}(\text{rank}(o \mid s,p)\leq k)\right),
\end{align*}
with $\mathds{1}(C)$ an indicator function that is 1 if the condition $C$ is true and $0$ otherwise.

\section{BioKG Node Degree Statistics\label{app:BioKGstats}}
\begin{table*}[ht]
\begin{center}
\centering
\begin{minipage}{\textwidth}
    \caption{Complete node degree statistics for BioKG}
    \begin{tabular}{lccccc}
        \toprule
         &  Mean degree & Median Degree & Degree std & Max Degree & Min degree \\\midrule
        \textbf{Total} & 39.19 & 5 & 171.07 & 2872 & 1\\
        DDI & 624.86 & 581.5 & 525.85 & 2477 & 1\\
        Protein Pathway Association & 8.46 & 3 & 20.80 & 1471 & 1\\
        PPI & 10.28 & 4 & 19.61 & 478 & 1\\
        Protein Disease Association & 7.72 & 2 & 38.11 & 2779 & 1\\
        Member of Complex & 8.21 & 3 & 17.95 & 421 & 1\\
        Drug Disease Association & 20.94 & 5 & 44.49 & 724 & 1\\
        DPI & 5.35 & 2 & 17.80 & 968 & 1\\
        Complex in Pathway & 2.88 & 1 & 6.43 & 158 & 1\\
        Complex Top Level Pathway & 2.40 & 1 & 41.58 & 2787 & 1\\
        Drug Target & 3.23 & 1 & 8.01 & 297 & 1\\
        Drug Pathway Association & 4.54 & 2 & 6.56 & 54 & 1\\
        Drug Enzyme & 4.85 & 2 & 27.67 & 958 & 1\\
        Disease Genetic Disorder & 3.02 & 1 & 6.26 & 46 & 1\\
        Related Genetic Disorder & 1.03 & 1 & 0.32 & 14 & 1\\
        Disease Pathway Association & 4.74 & 3 & 5.14 & 52 & 1\\
        Drug Transporter & 4.83 & 2 & 18.39 & 495 & 1\\
        Drug Carrier & 2.49 & 1 & 16.08 & 389 & 1\\\botrule
    \end{tabular}
    \label{tab:full_degree_stats}
\end{minipage}
\end{center}
\end{table*}
\clearpage

\section{Hyperparameters\label{app:hyperparameters}}
\begin{table*}[ht]
\begin{centering}
\begin{minipage}{\textwidth}
\caption{All LP hyperparameters and corresponding possible values.}
\begin{tabular}{@{}lc@{}}
\toprule
\multicolumn{1}{c}{Hyperparameter} & Values                                              \\ \midrule
Embedding size                             & {[}128, 256, 512, 1024{]}                   \\
Training type                      & [NegSamp, 1vsAll]                                   \\
\quad NegSamp neg. subjects              & {[}1, 100{]}                            \\
\quad NegSamp neg. objects               & {[}1, 100{]}                            \\
Max epochs                         & 200                                                 \\
Reciprocal                         & {[}True, False{]}                                   \\
Loss                               & CE                                                  \\
Optimiser                          & {[}Adam, Adagrad{]}                                 \\
\quad Batch size                             & {[}128, 256, 512, 1024{]}                 \\
\quad Learning rate                      & {[}\num{0.0003}, 1.0{]}                             \\
\quad Scheduler patience                 & {[}0, 10{]}                                   \\
L$_p$ regularisation          & {[}None, L1, F2, N3{]}                                   \\
\quad Entity emb. weight                 & $[1.0e-20, 1.0e-1]$                     \\
\quad Relation emb. weight               & $[1.0e-20, 1.0e-1]$                     \\
\quad Frequency weighting                & {[}True, False{]}                             \\
Dropout                            &                                                     \\
\quad Entity embedding                   & {[}-0.5, 0.5{]}                               \\
\quad Relation embedding                 & {[}-0.5, 0.5{]}                               \\
Embedding initialisation            & {[}Uniform, Normal, XavierUniform, XavierNormal{]} \\
\quad Normal mean                        & 0.0                                           \\
\quad Normal std.                        & {[}0.00001, 1.0{]}                            \\
\quad Uniform lower bound                & {[}-1.0, -0.00001{]}                          \\
\quad XavierUniform gain                 & 1.0                                           \\
\quad XavierNormal gain                  & 1.0                                           \\ \botrule
\end{tabular}
\end{minipage}
\end{centering}
\end{table*}
\clearpage

\section{Best-found configurations\label{app:config}}
\begin{table*}[ht]
\small
\caption{Hyperparameters and MRR of the LP models on BioKG as a whole. Best from 30 quasi-random HPO trials.}\label{tab:config}
\begin{tabular}{@{}lcccccc@{}}
\toprule
                              & \multicolumn{6}{c}{BioKG}                             \\
                              & ComplEx             & DistMult            & TransE                 & TransH   & RotatE & ConvE \\ \midrule
MRR (valid)                   & 0.630               & 0.165               & 0.274                  & 0.281    & 0.421  & 0.599     \\
MRR (test)                    & 0.629               & 0.471               & 0.273                  & 0.281    & 0.422  & 0.599    \\
Embedding size                & 512                 & 128                 & 256                    & 256      & 1024   & 1024     \\
Training type                 & 1vsAll              & 1vsAll              & 1vsAll                 & NegSamp  & NegSamp & 1vsAll   \\
\quad NegSamp neg. objects    & --                  & --                  & --                     & 51       & 1        & --       \\
\quad NegSamp neg. subjects   & --                  & --                  & --                     & 1        & 3        & --       \\
Epochs                        & 184                 & 14                  & 124                    & 200      & 200      & 94     \\
Reciprocal                    & Yes                 & No                  & Yes                    & Yes      & Yes      & Yes    \\
Loss                          & CE                  & CE                  & CE                     & CE       & CE       & CE     \\
Optimiser                     & Adag.               & Adag.               & Adag.                  & Adag.    & Adag.    & Adag.  \\
\quad Batch size              & 128                 & 128                 & 256                    & 256      & 128      & 512    \\
\quad Learning rate           & 0.417               & 0.417               & \num{9.85e-3}           & \num{8.61e-3} & \num{3.21e-3} & \num{2.02e-3}  \\
\quad Scheduler patience      & 10                  & 10                  & 10                     & 10       & 10       & 10       \\
L$_p$ regularisation          & F2                  & F2                  & None                   & None     & None     & None   \\
\quad Entity emb. weight      & \num{6.63e-7}        & \num{6.63e-7}        & --                     & --       & --       & --     \\
\quad Relation emb. weight    & \num{2.57e-15}       & \num{2.57e-15}       & --                     & --       & --       & --     \\
\quad Frequency weighting     & No                  & No                  & --                     & --       & --       & --     \\
Dropout                       &                     &                     &                        &          &          &        \\
\quad Entity embedding        & 0.407               & 0.407               & 0.117                  & 0.0208   & 0.000    & 0.000  \\
\quad Relation embedding      & 0.0370              & 0.0370              & 0.000                  & 0.000    & 0.0392   & 0.394  \\
Embedding initialisation      & XN($\sigma=0.0866$) & XN($\sigma=0.0866$) & N($\sigma=$\num{4.12e-4}) & N($\sigma=$\num{1.33e-3}) & XU($\pm 0.199$) & N($\sigma=$\num{1.26e-3}) \\ \botrule
\end{tabular}
\end{table*}

\begin{sidewaystable}[ht]
\footnotesize
\caption{Hyperparameters and MRR of ComplEx on the BioKG polypharmacy KGs. Best from 30 quasi-random HPO trials.}\label{tab:config-bench}
\begin{tabular}{@{}lcccccccc@{}}
\toprule
                              & \multicolumn{2}{c}{DDI-EFFICACY} & \multicolumn{2}{c}{DDI-MINERAL} & \multicolumn{2}{c}{DPI-FDA} & \multicolumn{2}{c}{DEP-FDA-EXP} \\
                              & ComplEx        & ComplEx-P        & ComplEx        & ComplEx-P        & ComplEx      & ComplEx-P     & ComplEx        & ComplEx-P        \\ \midrule
MRR (valid) & 0.838 & 0.859 & 0.861 & 0.885 & 0.383 & 0.540 & 0.171 & 0.184 \\
MRR (test)  & 0.847 & 0.865 & 0.861 & 0.884 & 0.386 & 0.542 & 0.171 & 0.185 \\
Embedding size & 512 & 512 & 512 & 512 & 512 & 512 & 512 & 512 \\
Training type & 1vsAll & 1vsAll & 1vsAll & 1vsAll & 1vsAll & 1vsAll & 1vsAll & 1vsAll \\
Epochs & 196 & 52 & 106 & 54 & 76 & 4 & 72 & 22 \\
Reciprocal & No & No & No & No & Yes & Yes & Yes & Yes\\
Loss & CE & CE & CE & CE & CE & CE & CE & CE \\
Optimiser & Adag. & Adag. & Adag. & Adag. & Adam & Adag. & Adag. & Adam \\
\quad Batch size & 128 & 256 & 128 & 256 & 128 & 1024 & 1024 & 1024 \\
\quad Learning rate & 0.0918 & 0.0229 & 0.0918 & 0.0170 & 0.0114 & \num{1.50e-3} & 0.156 & \num{5.81e-3} \\
\quad Scheduler patience & 10 & 10 & 10 & 10 & 10 & 10 & 10 & 10 \\
L$_p$ regularisation & N3 & N3 & N3 & L1 & None & None & None & None \\
\quad Entity emb. weight & \num{3.29e-18} & \num{9.36e-9} & \num{3.29e-18} & \num{3.15e-13} & -- & -- & -- & -- \\
\quad Relation emb. weight & \num{1.27e-5} & \num{3.73e-4} & \num{1.27e-5} & \num{2.03e-6} & -- & -- & -- & -- \\
\quad Frequency weighting & No & Yes & No & Yes & Yes & Yes & Yes & Yes \\
Dropout &&&&&&&&\\
\quad Entity embedding & 0.451 & 0.189 & 0.451 & 0.000 & 0.000 & 0.345 & 0.000 & 0.000 \\
\quad Relation embedding & 0.000 & 0.135 & 0.000 & 0.0863 & 0.159 & 0.462 & 0.259 & 0.0563 \\
Embedding initialisation & N($\sigma=$\num{1.63e-3}) & U($\pm0.700$) & N($\sigma=$\num{1.63e-3}) & U($\pm$\num{4.09e-4}) &  N($\sigma=$\num{1.78e-5}) & N($\sigma=$\num{3.02e-3}) & XN($\sigma=0.832$) & U($\pm0.329$) \\ \botrule

\end{tabular}
\end{sidewaystable}

\clearpage

\end{appendices}

\end{document}